\documentclass[conference]{IEEEtran}

\usepackage{graphicx}
\usepackage{times}
\usepackage{latexsym}
\usepackage{epsfig}
\usepackage{epstopdf}
\usepackage{tikz}
\usepackage{amssymb, amsmath}
\usepackage{amsfonts}
\usepackage{listings}
\usepackage{url}
\usepackage{mathtools}
\usepackage[
    bookmarks,
    bookmarksopen=true,
    colorlinks=true,
    linkcolor=black, 
    anchorcolor=black,
    citecolor=black, 
    filecolor=magenta,
    menucolor=red, 
    urlcolor=black,
    plainpages=false, 
    pdfpagelabels, 
    linktocpage,
    pdfpagemode={UseOutlines},
    bookmarksopenlevel=0,
    bookmarksnumbered,
    pdfstartview={FitV},
    unicode,
    breaklinks,
]{hyperref}

\begin{document}

\title{Reconstructing Neural Parameters and Synapses of arbitrary interconnected Neurons from their Simulated Spiking Activity}

\author{\IEEEauthorblockN{J\"orn Fischer}
\IEEEauthorblockA{
Mannheim University\\of Applied Sciences\\
Paul-Wittsack-Str. 10\\
68163 Mannheim\\
Germany\\
Tel.: (+49) 621 292-6767\\
Fax: (+49) 621 292-6-6767-1\\
Email: j.fischer@hs-mannheim.de\\
}
\and

\IEEEauthorblockN{Poramate Manoonpong}
\IEEEauthorblockA{
Embodied AI \& Neurorobotics Lab \\
Center for BioRobotics \\
The M{\ae}sk Mc-Kinney M{\o}ller Institute \\
University of Southern Denmark \\
DK-5230 Odense M \\
Denmark\\
Email: poma@mmmi.sdu.dk
}

\and
\IEEEauthorblockN{Steffen Lackner}
\IEEEauthorblockA{Mannheim University\\of Applied Sciences\\
Paul-Wittsack-Str. 10\\
68163 Mannheim\\
Email: steffen.lackner@posteo.de}
}

%

\date{Received: 2016-08 / Accepted: 2016-08}

\maketitle

\begin{abstract}

To understand the behavior of a neural circuit it is a presupposition that we have a model of the dynamical system describing this circuit. This model is determined by several parameters, including not only the synaptic weights, but also the parameters of each neuron. Existing works mainly concentrate on either the synaptic weights or the neural parameters. 
In this paper we present an algorithm to reconstruct all parameters including the synaptic
weights of a spiking neuron model. The model based on works of Eugene M. Izhikevich \cite{izhikevich2007dynamical} consists of two differential equations and covers different types of cortical neurons. It combines the dynamical properties of Hodgkin-Huxley-type dynamics with a high computational efficiency. The presented algorithm uses the recordings of the corresponding membrane potentials of the model for the reconstruction and consists of two main components. The first component is a rank based Genetic Algorithm (GA) which is used to find the neural parameters of the model. The second one is a Least Mean Squares approach which computes the synaptic weights of all interconnected neurons by minimizing the squared error between the calculated and the measured membrane potentials for each timestep.
In preparation for the reconstruction of the neural parameters and of the synaptic weights from real measured membrane potentials, promising results based on simulated data generated with a randomly parametrized Izhikevich model are presented. The reconstruction does not only converge to a global minimum of neural parameters, but also approximates the synaptic weights with high precision.

\end{abstract}

\section{Introduction}

The strength of synaptic connections as well as the cell parameters of spiking networks or cell assemblies \cite{fischer} are key ingredients underlying higher order cognitive functions \cite{izhikevich2007dynamical}. According to this, there are numerous works which have developed techniques to obtain the network parameters based on physiological and anatomical measured data \cite{hatsopoulos2009science,ghane2013design,oh2014mesoscale,grewe2009optical,lutcke2013inference,olsen2008cracking}.

One could try to get the interconnection matrix via parallel recordings of their spiking activity. Interesting techniques are e.g. the multi-electrode array technology for in vivo implantation (\cite{hatsopoulos2009science} and \cite{ghane2013design}), which has been described as a method to be used to implement Neural Interface Systems also known as Brain-Machine interfaces. Nevertheless it may also be useful for recording spiking activity in general. The resulting data then can be used for the reconstruction of interconnections between up to thousands of nerve cells.

Progress is also made in techniques to investigate larger networks. These include macroscale, microscale, and meso\-scale connectivity mapping techniques \cite{oh2014mesoscale}. On microscale, on which also individual synapses can be ana\-lysed, only a limited number of cells can be observed due to the high effort this method requires. The mesoscale techniques on the other hand are insufficient at the moment to deliver the true functional connection of the synapses.

Other methods are e.g. optical methods, as described in \cite{grewe2009optical}, \cite{lutcke2013inference}, and also \cite{olsen2008cracking}. The latter for example describes imaging methods based on the analysis of green fluorescent proteins (GFP) in the neurons as well as transmembrane carrier proteins, and makes it possible to determine if multiple neurons are synaptic connected.

There are many types of neural spiking models on different levels of complexity. On the one hand simple reduced models are often unable to resolve measured scenarios, on the other hand models consisting of many non-linear differential equations are mathematically hard to handle. One of the models which are more difficult to be calculated is the Hodgkin-Huxley model \cite{hodgkin1952quantitative}, in which the ion-channel currents of the real neuron are simulated. There are also simpler models which still provide a rich spiking and bursting dynamics, as e.g. the \cite{izhikevich03simplemodel} model which we will use in this paper. A more detailed comparison of the models is also made in \cite{izhikevich2004model}, where Eugene M. Izhikevich compares also the computational complexity.

In \cite{kumar2010optimal} an estimation of the parameters of the Izhikevich model is made using the data of inter-spike intervals for a single neuron with experimental data from a primate study. However, in contrast to the approach of Kumar et al, we use the full measured membrane data and aim to reconstruct a whole network instead of a single neuron.

Tokuda et al also proposed an estimation of the parameters for the Hindmarsh-Rose model \cite{hindmarsh1984model}. The disadvantage of this model is that it is also more complex and computational expensive compared to the model of Eugene M. Izhikevich.


\cite{zaytsev2015reconstruction} introduced a new reconstruction method
based on Maximum Likelihood Estimation for a generalised
one dimensional neuron model. Though this method
is the most promising one for synaptic weight reconstruction, it shows several
weaknesses, which can be overcome by our new approach.

The weaknesses can be summarized as follows:
\begin{itemize}
        \item The model is based on an inhomogeneous Poisson process, which is not able to model a recovery-variable or a timing dependent distribution. As an example, a spike train which is chattering is not covered by this model. 
        \item The recovery of all neural parameters, which are necessary to build a complete model of the neural network, is missing.
        \item Only the spiking information is used, not the information covered in the whole measured trajectory.
        \item The recurrency of the neuron itself, as well as the information of an external input is treated differently from that of the signal of other neurons.
\end{itemize}


The next section gives a short description of the Eugene M. Izhikevich neural
model used in this study. The sections 3,4 describe our new approach which reconstructs a complete spiking neural model from the membrane
potentials of a fixed number of interacting neurons. The section 5 provides experimental results while the section 6 concludes the study.

\section{Spiking neuron model}
To simulate a network of spiking neurons we use the Izhikevich model \cite{izhikevich03simplemodel}. 
It is based on a system of two differential equations, which describe a fast voltage variable and a slow recovery variable.
The model is computational inexpensive compared to the Hodgkin-Huxley model but still provides rich spiking and bursting dynamics. It describes a two dimensional non linear system of coupled differential equations containing four dimensionless control parameters $a_{i}$, $b_{i}$, $c_{i}$, and $d_{i}$, which govern the dynamics.
\begin{align}
        \frac{dv_{i}}{dt} = 0.04v_{i}^{2} + 5v_{i} + 140 - u_{i} + I_{i} \\
        \frac{du_{i}}{dt} = a_{i}(b_{i}v_{i} - u_{i})\\
        \text{with} \hspace{0.2cm} I_{i} = \sum_{i}{v_{j} w_{ij}} 
\end{align}

The model contains a recovery sequence when a neuron gets fired:
\begin{align}
        if (v_{i} >= 30)
        \begin{cases}
                v_{i} \leftarrow c_{i}, u_{i} = u_{i} + d_{i},
        \end{cases} 
        \label{recoverySequence}
\end{align}

where $v_{i}$ is the membrane potential of neuron $i$. The current $I_{i}$ is the input current which is given if a connected neuron releases an action potential one time step before. The variable $u_{i}$ is the membrane recovery variable with the intention to simulate the repolarisation and $c_{i}$ is the value on which the membrane potential is set after an action potential occurred.

The variables $a_{i}$, $b_{i}$, and $d_{i}$ affect the membrane recovery variable $u_{i}$; where $a_{i}$ affects the recovery speed of $u_{i}$ (higher values mean fast recovery, and vice versa), $b_{i}$ determines how strong $u_{i}$ and $v_{i}$ are coupled together, and $d_{i}$ is responsible for the after-spike reset difference of $u_{i}$. Figure. \ref{fig:izhikevichFig} shows how the parameters $a_{i}$, $b_{i}$, $c_{i}$, and $d_{i}$ are chosen to get a specific type of cortical neuron (in analogy to \cite{izhikevich03simplemodel}).
To keep things simple, we numerically compute the differential equations in small time steps of $\Delta t=0.5 ms$ using the Euler method. This leads to the following equations: 

\begin{figure*}[htb]
        \centering
        \includegraphics[scale=0.65]{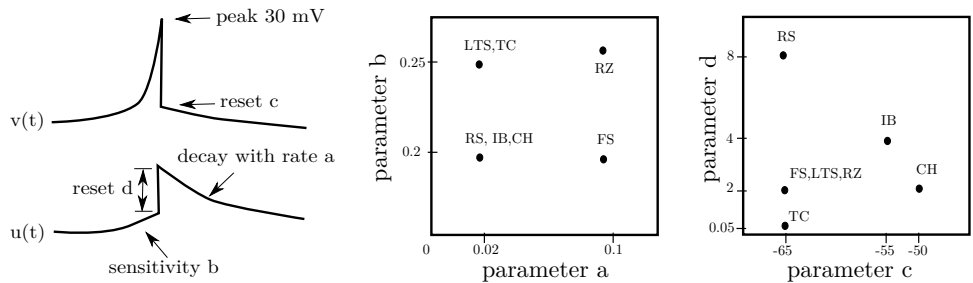}
        \caption{Different types of cortical neurons correspond to different parameters $a_{i}$, $b_{i}$, $c_{i}$, and $d_{i}$. Eugene M. Izhikevich shows that the plotted parameters are related to regular spiking (RS), intrinsically bursting (IB), chattering (CH), fast spiking (FS), thalamo-cortical (TC), resonator (RZ), and low-threshold spiking (LTS) neurons (Original from \cite{izhikevich03simplemodel}, Created with SciDAVis: \url{http://scidavis.sourceforge.net/})}
        \label{fig:izhikevichFig}
\end{figure*}

\begin{align}
        v^{t+1}_{i} = v^{t}_{i}+\Delta t(0.04(v^{t}_{i})^{2} + 5v^{t}_{i} + 140 - u^{t}_{i} + I_{i}) \label{membPot}\\
        u^{t+1}_{i} = u^{t}_{i}+\Delta t(a_{i}(b_{i}v^{t}_{i} - u^{t}_{i})) \label{recovery}\\
        \text{with} \hspace{0.2cm} I_{i} = \sum_{i}{v^{t}_{j} w_{ij}} \label{inputEquation}
\end{align}

The recovery sequence stays the same as described in equation \ref{recoverySequence}.

\section{Genetic Algorithm (GA) for parameter estimation}
As a preparation to reconstruct the synaptic weights, the parameters of the model $a_{i}$, $b_{i}$, $c_{i}$, $d_{i}$, and $u_{i}^{t=0}$ have to be chosen. 

One might think that the parameter $c_{i}$ could be directly read from the data by analysing the membrane potential value immediately after a spike. This is not the case because the weakness of the Izhikevich model is the after spike reset which is an artificial and discontinuous behaviour. Though the membrane potential falls to a minimum in a very short time, in general it needs more than half a millisecond. The determination of the best fitting reset value from real measured data is more complex and can be solved treating $c_{i}$ as a usual parameter of the reconstruction algorithm.

\begin{figure}[htb]
        \begin{center}
                \includegraphics[scale=0.3]{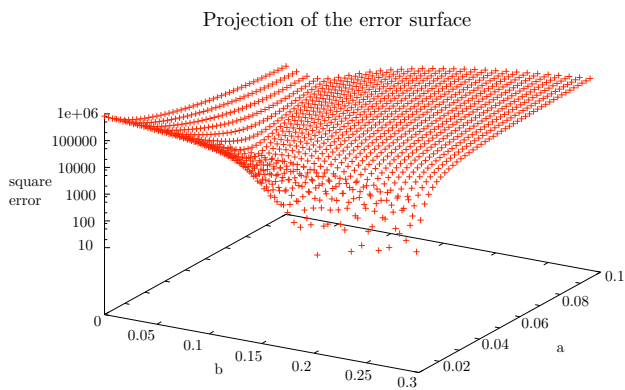}
                \caption{3D plot of the squared error in dependence of parameter $a_{i}$ and $b_{i}$. The Parameters $c_{i}$, $d_{i}$, and $u_{i}^{t=0}$ are still optimal. The projection of the error surface is smooth, so that a Genetic Algorithm might be well suited to reach the global minimum (Created with SciDAVis)}
                \label{fig:error}
        \end{center}
\end{figure}

To have an idea about the smoothness of the error surface the mean squared error is plotted in a projection, which shows the error of the model with respect to the variables $a_{i}$ and $b_{i}$ as shown in figure \ref{fig:error}.

Because a gradient may not be computed easily we cho\-ose a Genetic Algorithm with a rank based selection for parameter estimation. The best individual is always selected, and protected against mutation. A population size of $p=1000$ individuals shows good results, and nearly always rea\-ches the global optimum in only a few hundred generations. The used mutation operator just changes 1 Bit of the parameter set. Each parameter is implemented as a 16 bit integer number, transformed from genotype to phenotype by scaling and shifting to fit into its predefined value range.
The crossover operator is implemented with a single crossover point between two parameters. The crossover rate as well as the mutation rate are set to $r=m=0.5$. The mutation operator is implemented by transforming the floating point parameter into the corresponding 16 bit integer number. This number is then transformed into Gray code, to ensure that neighboured states only differ in one bit. Afterwards a random bit is toggled and the Gray code transformed again into an integer. Finally it is converted into the corresponding floating point parameter. 

On the basis of the estimated parameters the reconstruction of the synaptic weights is made, which is described in the following section.

\section{Synaptic weight reconstruction}
In this section, we describe the algorithm to reconstruct the synaptic weights which is based on the equations \ref{recoverySequence} to \ref{inputEquation}. At first we assume that the weights in the measured time frame are constant.
For this paper the Least Squares Method is hereby used to find a solution which minimizes the squared error to reconstruct the synaptic weights.
We assume that the membrane potential of each neuron is known for at least $n+1$ time steps, where $n$ is the number of interconnected neurons.

\tikzstyle{every picture}+=[remember picture]
\everymath{\displaystyle}

To reconstruct the synaptic weights the equations have to be transformed into an optimization problem.
The error is the difference between the calculated \tikz\node[fill=blue!18,dotted,draw,thin]  (n2) {}; and the measured \tikz\node[fill=red!18,dashed,draw,thin] (n1) {}; membrane potential.

{\small
\begin{equation}{\scriptstyle
        E = 
        \tikz[baseline] {
                \node[fill=blue!5,anchor=base,dotted,draw,thin] (t1)
                {$ (v_i^{t} + \Delta t  (0.04 (v_{i}^{t})^{2} + 5 v_{i}^{t} + 140 - u_{i}^{t} + I_i ) $};
        }
        - \tikz[baseline]{ 
                \node[fill=red!5,anchor=base,dashed,draw,thin] (t2)
                {$ (v_{i}^{t+1}) $};
        }
        }
        \label{eq:error}
\end{equation}}

To minimize the squared error, we calculate the derivative with respect to the weights and set it to zero:
\begin{equation}
        \frac{\partial E^2}{\partial W_{ij}} =  0 \label{equSys},
\end{equation}

where index j lies in the interval $[1;n]$ and  where $n$ is the amount of neurons of the network. In other words, equation \ref{equSys} forms a system of linear equations with $j$ as an index for the $j$-th equation. Dissolving the derivative we obtain the equation \ref{eq:last}.

\begin{equation}
    \splitfrac{\sum_{t} \sum_{i} 2v_{j}^{t+1} (v_{i}^{t+1} - (v_{i}^{t} + h(0.04 (v_{i}^{t})^{2} +  5 v_{i}^{t} }{+ 140 - u_{i}^{t}) + \sum_{i}{v^{t}_{j} w_{ij}}) = 0}
        \label{eq:last}
\end{equation}

This equation system must be solved independently for each neuron of the network.
Since afterwards all information of the network is given, the error of the network for the parameters $a_{i}$, $b_{i}$, $c_{i}$, $d_{i}$, and $u_{i}^{t=0}$ can be calculated easily using the equation \ref{equSys}. This error can be used as an inverse fitness measure, which describes how good the parameters $a_{i}$, $b_{i}$, $c_{i}$, $d_{i}$, and $u_{i}^{t=0}$ are chosen, and is used in the GA of the proposed solution.

It is important that the spike data is not included in the equation system because there are points of discontinuity. This is because of the recovery sequence, which is implemented as an if-condition described in the equation \ref{recoverySequence}.

\section{Experimental results}
We performed several experiments to evaluate the performance of the proposed approach. In all experiments, we set the four control parameter as $a_{i}=0.02$, $b_{i}=0.2$, $c_{i}=-55$, $d_{i}=4$, and $u_{i}^{t=0}=-11$. The type of neuron corresponding to the values of $a_{i}$, $b_{i}$, $c_{i}$, and $d_{i}$ is intrinsically bursting, see figure \ref{fig:izhikevichFig}. Different values for the control parameters correspond to other known types. 

The initial value of the membrane potential $v_{i}^{t=0}$ is initialized with the after spike reset parameter $c$. The potential $v_{i}$ always lies in the interval $[-65, 30]$. To limit each parameter on its natural interval, we use the description of \cite{izhikevich03simplemodel} defining the intervals as:

\begin{equation}
a \in [0.01, 0.1], b \in [0.05, 0.3], c \in [-65, -50], d \in [0.05, 8]
\label{lineup1}
\end{equation}

The interval for $u_{i}$ has not been described in \cite{izhikevich03simplemodel}. However, experiments have shown that the parameter $u_{i}$ is limited to:

\begin{equation}
u^{t=0} \in [-15, 15].
\label{uInterval}
\end{equation}

Using the predefined parameter set $a_{i}=0.02$, $b_{i}=0.2$, $c_{i}=-55$, $d_{i}=4$, and $u_{i}^{t=0}=-11$, it is shown that the weight matrix can be reconstructed with an accuracy of four decimal places.

\begin{figure}[htb]
        \begin{center}
                \includegraphics[scale=0.23]{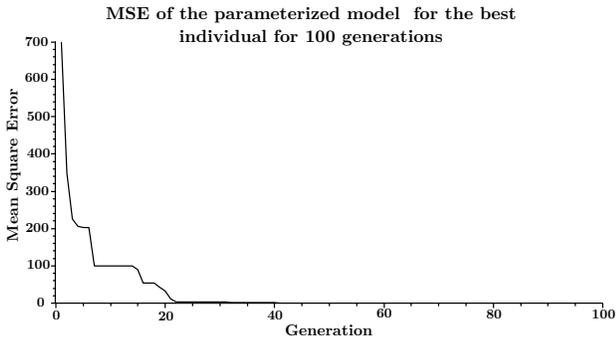}%
        \caption{The resulting Mean Square Error during a reconstruction. It was observed that after about 35 generations, the individuals show only small improvements and all parameters were already close to the optimum (Created with SciDAVis)}
                \label{fig:reconstructionFitness}
        \end{center}
\end{figure}

Applying the Genetic Algorithm with rank selection, we only need about $100$ generations with a population size of $1000$ individuals to reach a nearly optimal parameter set. The figure \ref{fig:reconstructionFitness} shows the resulting Mean Square Error during a reconstruction process. 

\begin{figure*}
        \begin{center}
                \includegraphics[scale=0.50]{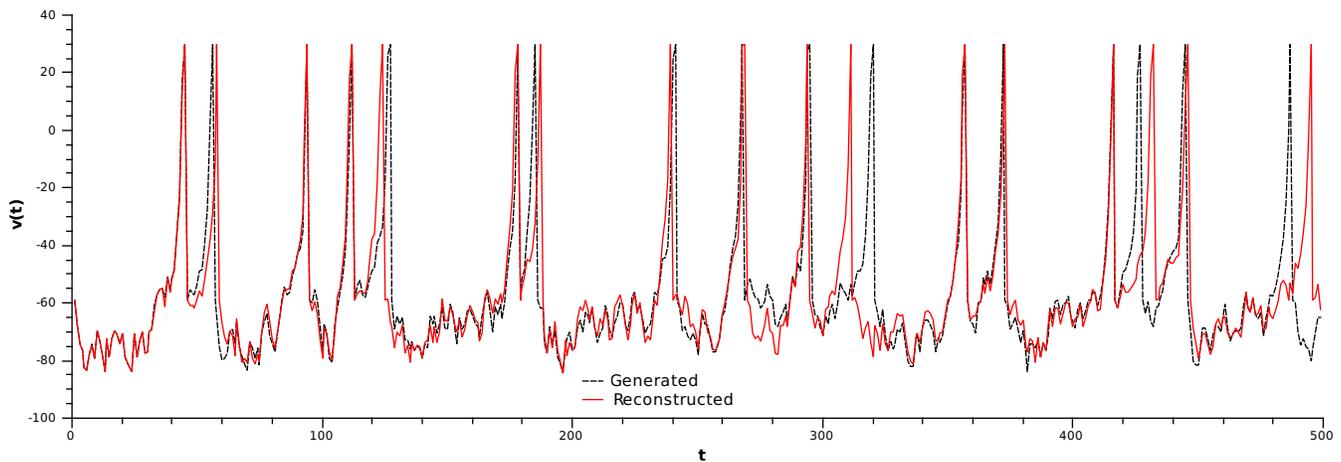}%
        \caption{The result of a reconstruction of 10 neurons. The reconstructed as well as the generated data of the first neuron is shown in this chart. Though the starting parameters are nearly identical, the trajectory of both neurons are slightly diverging. The qualitative behaviour defined by which type of neuron it is and which weights are reconstructed is in both cases the same (Created with SciDAVis)}
                \label{fig:reconstructionResult}
        \end{center}
\end{figure*}
\begin{figure}[htb]
    \begin{center}
        \includegraphics[scale=0.35]{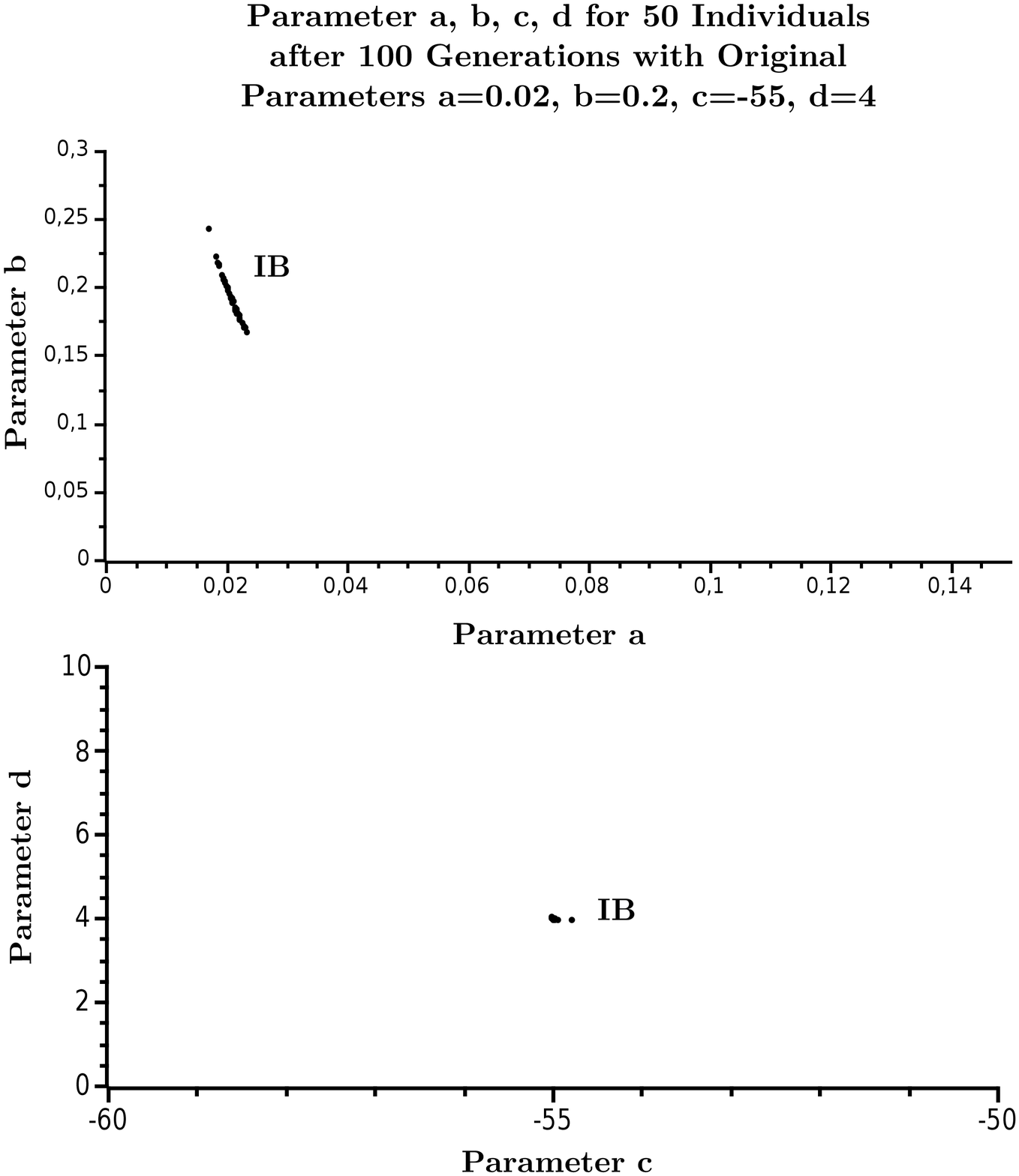}%
        \caption{The estimated parameters of $50$ individuals after $100$ Generations of Evolution. Though Parameter $a$ and $b$ are not perfectly approximated, they all seem to converge to the same original parameter set, the global optimum (Created with SciDAVis)}
        \label{fig:reconstructionParameters}
    \end{center}
\end{figure}

In figure \ref{fig:reconstructionResult} a comparison of a generated neuron and the corresponding reconstructed neuron is shown. We can see that though starting with a very similar parameter set the trajectory of the reconstructed neuron slightly diverges from the original trajectory. The qualitative behaviour defined by which type of neuron it is and which weights are reconstructed is in both cases the same.

\begin{figure*}[htb]
    \begin{center}
        \includegraphics[scale=0.5]{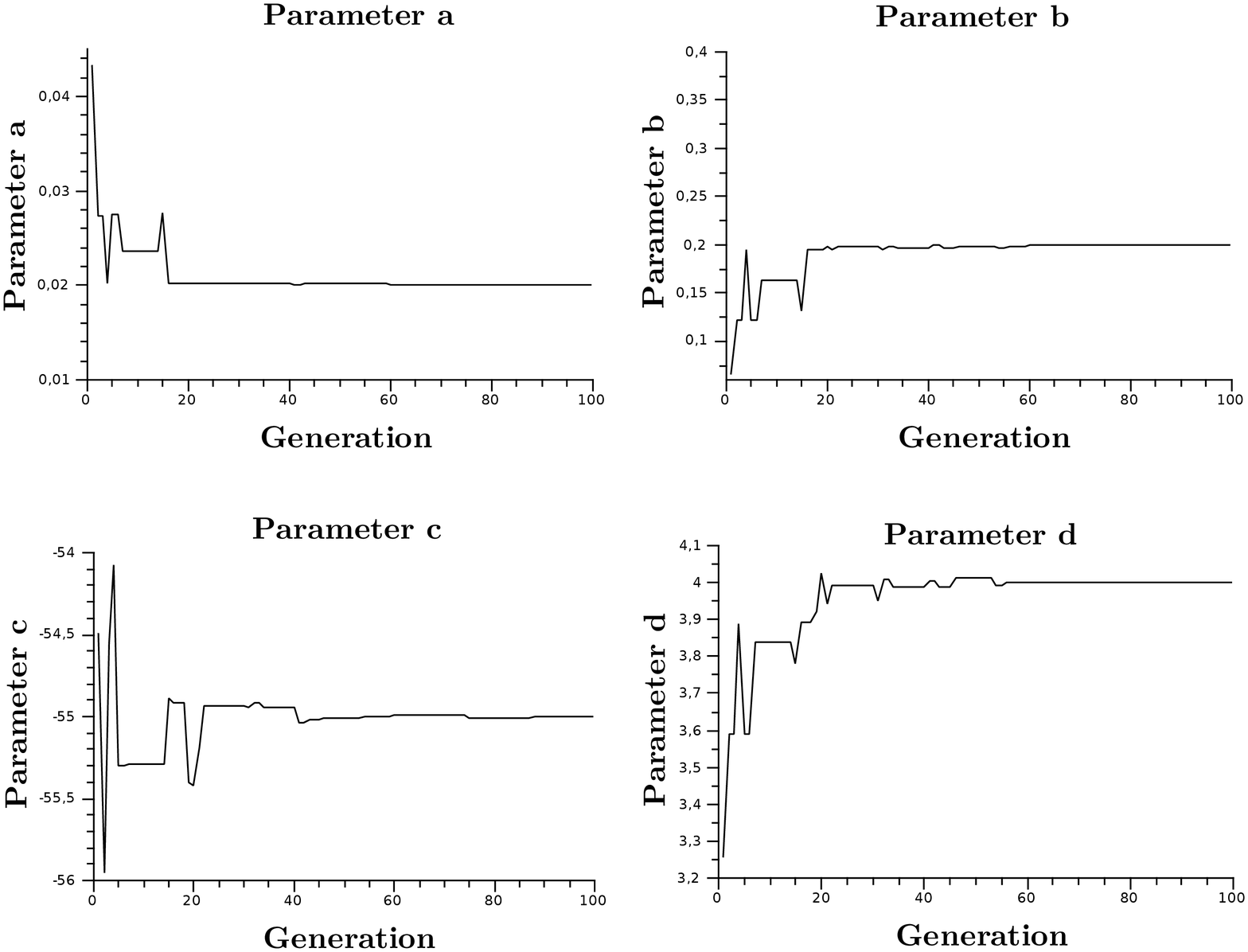}%
        \caption{As an example the parameters $a$, $b$, $c$, and $d$ of the best individual converge after about $60$ generations (Created with SciDAVis)}
                          \label{fig:parametersAtoD}
    \end{center}
\end{figure*}

\begin{figure*}[htb]
    \begin{center}
        \includegraphics[scale=0.5]{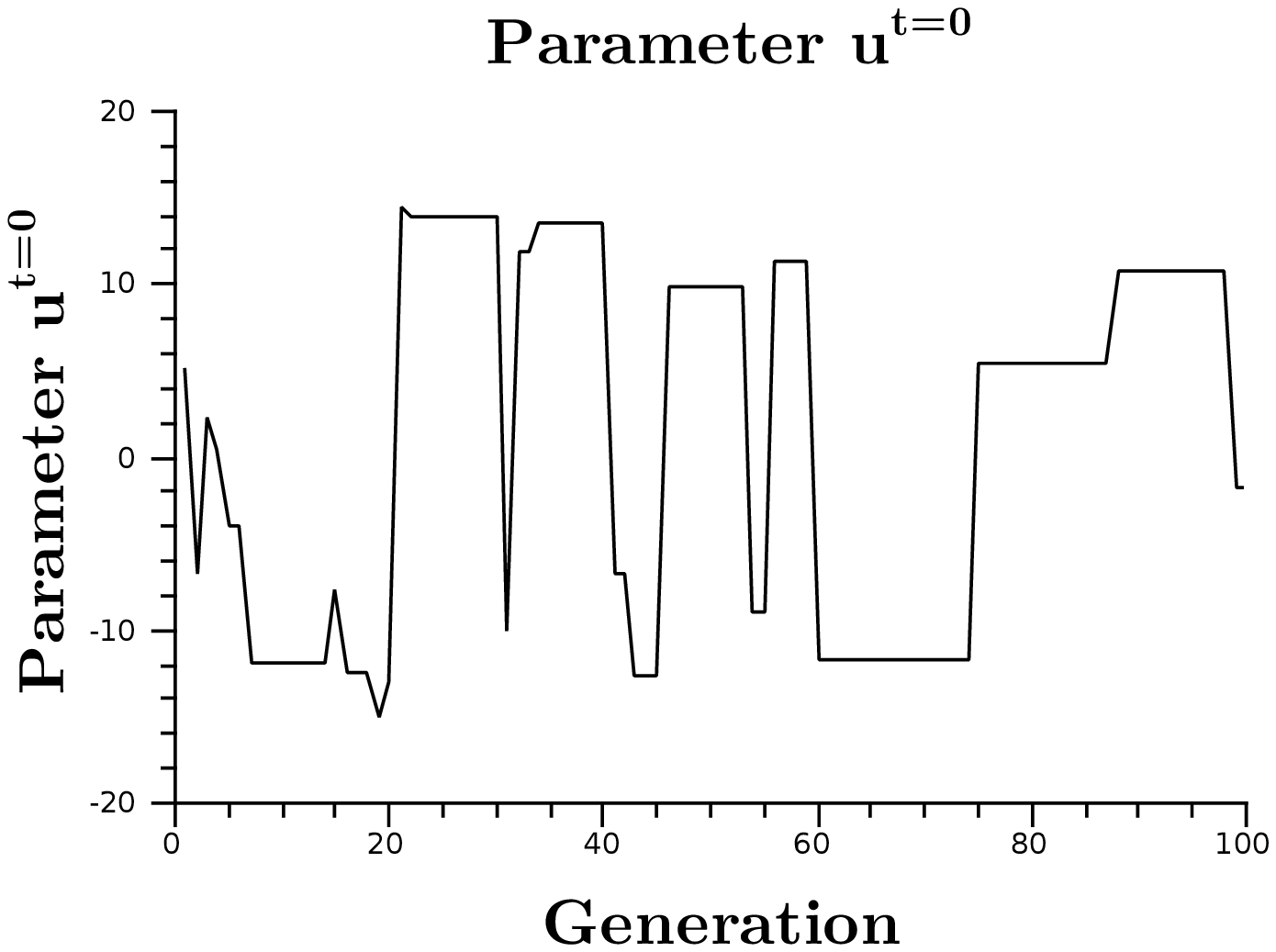}%
        \caption{In the same example the parameter $u_{i}^{t=0}$ of the best individual does not converge at all. The mean squared error seems to be widely independent of this initial value. Our suggestion is that the value of $u_{i}$ will swing into an attractor independent from its initial value (Created with SciDAVis)}
        \label{fig:parametersU}
    \end{center}
\end{figure*}

Figure \ref{fig:reconstructionParameters} and figure \ref{fig:parametersAtoD} shows that during the evolution the parameters $a$, $b$, $c$, and $d$ visibly converge after about $60$ generations, while in figure \ref{fig:parametersU} we can see that the mean squared error seems to be widely independent on the parameter $u_{i}^{t=0}$ which does not converge at all. Our suggestion is that the value of $u_{i}$ will swing into an attractor independent from its initial value. A strategy then could be to initialize $u_{i}^{t=0}=0$ and to wait a few time steps until $u_{i}$ has reached this attractor before using the data to calculate the weights. The Genetic Algorithm then would only have to optimize the four parameters $a$, $b$, $c$, and $d$.

The parameter search and the weight reconstruction is applied to each of the neurons independently. The computation time rises with $O(n^4)$ because a system of n linear equations has to be solved for each neuron to reconstruct the complete weight matrix. To be precise enough, so that simulation results are similar to the corresponding measured scenario, the precision of the parameters must be very high. We have to evolve several hundred of generations to get a sufficient precision. Otherwise for a large number of neurons the summation of all errors may lead to a diverging behaviour.

\section{Conclusion}
In this paper a feasibility study is made to check if it is possible to reconstruct the synaptic weights as well as all cell parameters of a simulated network of cortical spiking neurons by only using their membrane potential.

The greatest benefit of the developed method is, that a reconstruction of the complete model only needs the measured membrane potential for at least $n+1$ time steps, where $n$ denotes the number of neurons. The reconstructed network should hereafter behave similar to the real one, imitating the neurons spiking behaviour. However, it is likely that using real neuron data will lead to much more difficulties for the reconstruction algorithm, because also delay times between neurons, measuring errors and other circumstances have to be taken into consideration.
On the other hand, the model of Eugene M. Izhikevich offers a quite good approximation of the real neural dynamics, so that all the complex processes including neurotransmitters, ion-channel currents, etc. can be neglected.

Compared with other approaches like the one of Zaytsev, Morrison, and Deger our method also covers a recovery variable, which allows e.g. bursting dynamics, which is not possible in the model described in \cite{zaytsev2015reconstruction}. Furthermore, our approach delivers five parameters, which determine the complete model of the neuron. This leads to a reconstruction mechanism which could imitate the real neural spiking behaviour in a more realistic manner, because the Izhikevich model is able to simulate many different kinds of neuron types from biological brains (see also figure \ref{fig:izhikevichFig}).

In summary, by using a combination of the Least Mean Squares method to reconstruct the synaptic weights and the Genetic Algorithm for the cell parameters, it is shown that for the generated data an efficient reconstruction of the original parameters is possible. In future research this method, with promising results on generated data, is also tested on measured data of real neurons.


\section*{Acknowledgment}
The authors would like to thank Deniel Horvatic and Andreas Knecht for their helpful contributions.

\section*{Appendix}
The Java implementation which was used in our approach, as well as the generated data are available online:\\
\url{http://services.informatik.hs-mannheim.de/~fischer/publikationen.html}

\bibliographystyle{IEEEtran}
\bibliography{ReconstructionNeuralParameters_jFischer_pManoonpong_sLackner}

\end{document}